\documentclass[letterpaper]{article} 
\usepackage{aaai24}  
\usepackage{times}  
\usepackage{helvet}  
\usepackage{courier}  
\usepackage[hyphens]{url}  
\usepackage{graphicx} 
\usepackage{natbib}  
\usepackage{caption} 
\usepackage{bbding}
\usepackage{booktabs}
\usepackage{multirow}
\usepackage{algorithm}
\usepackage{algorithmic}
\usepackage{amsmath}
\usepackage{amssymb}
\usepackage{newfloat}
\usepackage{listings}
\usepackage{bibentry}
\usepackage{makecell}
\usepackage{xspace}
\usepackage{xcolor}
\usepackage{cleveref}

\crefname{section}{Sec.}{Secs.}
\Crefname{section}{Section}{Sections}
\crefname{table}{Tab.}{Tabs.}
\Crefname{table}{Table}{Tables}
\crefname{figure}{Fig.}{Figs.}
\Crefname{figure}{Figure}{Figures}
\crefname{equation}{Eq.}{Eqs.}
\Crefname{equation}{Equation}{Equations}

\DeclareCaptionStyle{ruled}{labelfont=normalfont,labelsep=colon,strut=off} 
\floatstyle{ruled}
\newfloat{listing}{tb}{lst}{}
\floatname{listing}{Listing}
\newcommand{\method}{AE-NeRF\xspace}
\newcommand{\ar}{audio related\xspace}
\newcommand{\ai}{audio independent\xspace}
\newcommand{\aac}{audio associated\xspace}
\newcommand{\ad}{audio decoupled\xspace}
\newcommand{\rrs}{Regionwise Ray Sampling\xspace}
\newcommand{\aaa}{Audio Aware Aggregation\xspace}
\newcommand{\afag}{Audio-Aligned Face Generation\xspace}

\makeatletter
\renewcommand{\maketag@@@}[1]{\hbox{\m@th\normalsize\normalfont#1}}%
\makeatother
 
\title{\method: Audio Enhanced Neural Radiance Field for Few Shot \\Talking Head Synthesis }
\author{
    Dongze Li\textsuperscript{\rm 1,2}, Kang Zhao\textsuperscript{\rm 3}, 
    Wei Wang\textsuperscript{\rm 2}\thanks{ Corresponding author.}, Bo Peng\textsuperscript{\rm 2}, \\
    Yingya Zhang\textsuperscript{\rm 3}, Jing Dong\textsuperscript{\rm 2}, Tieniu Tan\textsuperscript{\rm 2,4}
}
\affiliations{
    \textsuperscript{\rm 1}School of Artificial Intelligence, University of Chinese Academy of Sciences\\
    \textsuperscript{\rm 2}CRIPAC \& MAIS, Institute of Automation, Chinese Academy of Sciences\\
    \textsuperscript{\rm 3}Alibaba Group \\
    \textsuperscript{\rm 4}Nanjing University
}

\usepackage{bibentry}

\begin{document}
\maketitle
\begin{abstract}
Audio-driven talking head synthesis is a promising topic with wide applications in digital human, film making and virtual reality. 
Recent NeRF-based approaches have shown superiority in quality and fidelity compared to previous studies.  
However, when it comes to few-shot talking head generation, a practical scenario where only few seconds of talking video is available for one identity, two limitations emerge: 1) they either have no base model, which serves as a facial prior for fast convergence, or ignore the importance of audio when building the prior; 2) most of them overlook the degree of correlation between different face regions and audio, e.g., mouth is audio related, while ear is audio independent.
%
   In this paper, 
   we present Audio Enhanced Neural Radiance Field (\method) to tackle the above issues,
   which can generate realistic portraits of a new speaker with few-shot dataset.
   Specifically, we introduce an \aaa module into the feature fusion stage of the reference scheme, where the weight is determined by the similarity of audio between reference and target image.
   Then, an \afag strategy is proposed to model the audio related and audio independent regions respectively, with a dual-NeRF framework.
   Extensive experiments have shown \method surpasses the state-of-the-art on image fidelity, audio-lip synchronization, and generalization ability, even in limited training set or training iterations.
\end{abstract}

\section{Introduction}
 Audio-driven talking head generation is an essential technique with broad application scenarios such as digital human, film making, video conference and virtual reality. 
Many literature \cite{wav2lip,shen2023difftalk,ye2023geneface} have been put forward to learn the audio-to-lip mapping by using deep generative models, such as GAN, diffusion model, VAE, etc.
Among them, Neural Radiance Field (NeRF) \cite{mildenhall2020nerf} based methods \cite{adnerf,sspnerf,dfrf} have shown promising results, which map audio features to dynamic neural radiance fields to model a talking head.

However, NeRF-based methods usually adopt identity-specific training, i.e., one needs to train a model from scratch for each new identity. What's worse, to make model generalize to various mouth shapes, the training set for each identity should be large, which is difficult to be satisfied in practice since the data for one identity are often limited. One-shot talking head generation \cite{atvg,wav2lip,makeittalk} may be a solution, which 
drives the novel identity from one reference image without training.
But it sacrifices the fidelity of talking head, especially the teeth consistency and details. To balance data availability, training efficiency and generation quality, we focus on a practical scenario: few-shot talking head synthesis, that is, we need to train a NeRF model rapidly on a short talking video of one identity, which is capable of generating high-fidelity talking head with a given audio. Existing NeRF-based methods suffer from the following limitations when applied to this setting:

\begin{figure}[t]
\begin{center}
\includegraphics[width=0.99\linewidth]{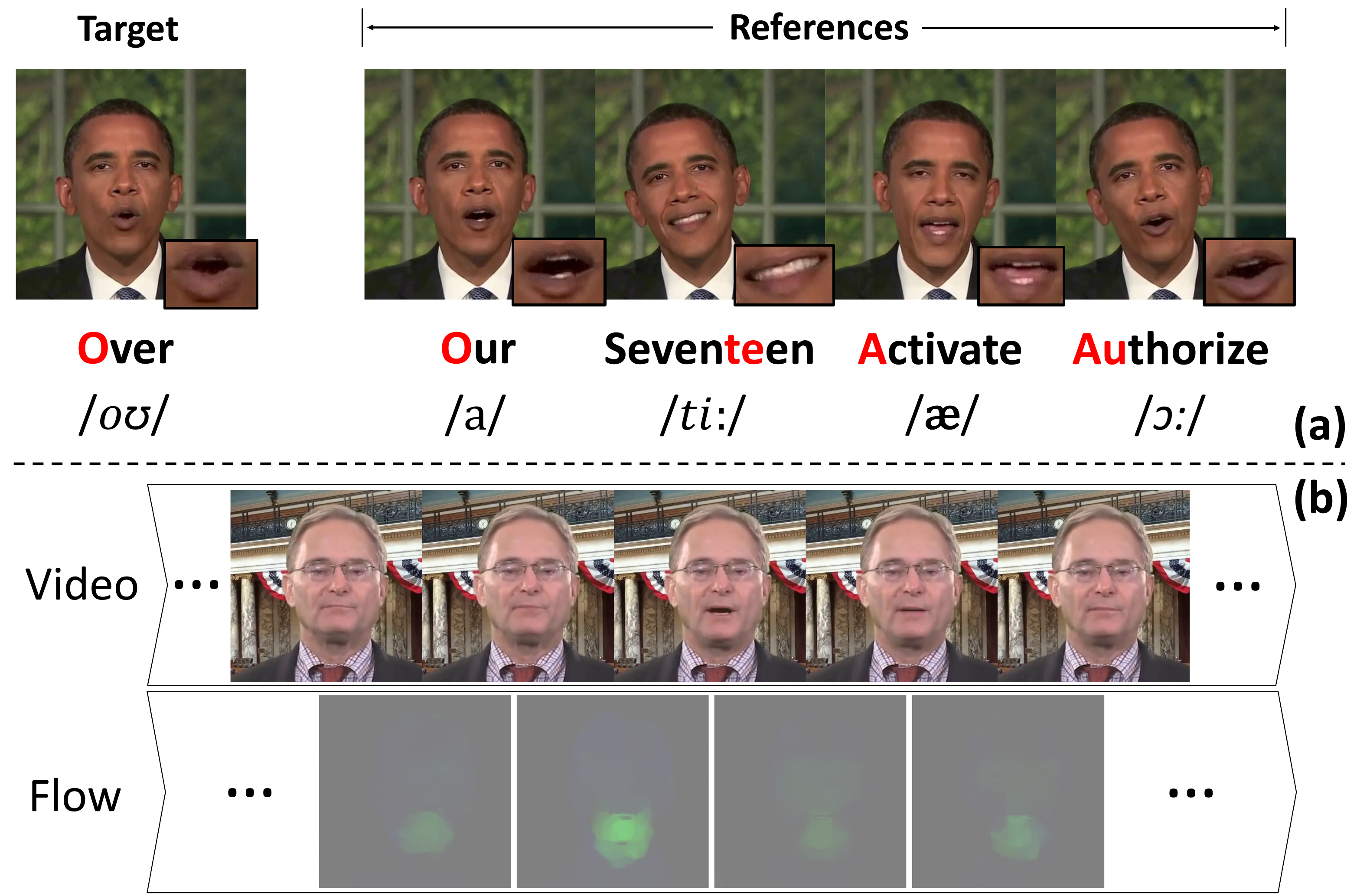}
\end{center}
   \caption{Our observation. (a) It shows the target/reference images and their phonetic symbols. The target is more similar to the first and last reference image because their pronunciation is closer. (b) We calculate the flow of adjacent frames, it can be seen the lower half face varies obviously than other regions.}
\label{fig_mot}
\end{figure}

\noindent\textbf{Lack of a robust prior.} 
In order to quickly generalize to the few-shot identity, it is necessary to pre-train a base model across multi-identity to provide a basic audio-to-lip translation and implicit facial priors, such as color, shape and texture, which are helpful to restore faithful facial details. Current methods either lack the prior or construct a less robust prior. The former \cite{adnerf,sspnerf} is trained directly on the dataset of one identity. When the data is limited, the accuracy of rendering will drop significantly. The latter \cite{dfrf} ignores the correlation of audio between the target and reference images.


\noindent\textbf{Audio-Face misalignment.} 
In the audio-driven neural radiance field, most methods render the color of each ray conditioned on the audio feature, which means the entire talking head is considered to be \textit{audio related}. Actually, we empirically find there exist many \textit{audio decoupled} regions that have a weak or no correlation with the audio signal, such as hairs, ears and wrinkles. Modeling these two regions (i.e., audio related and decoupled) with one radiance field will cause a misalignment between the visual and audio information, resulting in sub-optimal synthesis results.



In this work, we propose Audio Enhanced NeRF (AE-NeRF) to tackle these two issues. Based on a reference scheme, we introduce \textit{\aaa} module and \textit{\afag} strategy, to empower NeRF models with the ability to synthesize high quality talking heads with limited training data.

Specifically, we pre-train a base model on multi-identity dataset first. For each identity, we input several reference images to provide visual information from different poses to help the model render the target image. We find that \textit{the close pronunciations have the similar mouth shapes} (see (a) of \cref{fig_mot}). Therefore, when aggregating the visual features from the reference images, whose weight should be higher if its audio is closer to that of the target image. For an audio-driven application, the \textit{\aaa} module will make the learned prior more robust and accurate.

\begin{figure*}[t]
\centering
\includegraphics[width=1\linewidth]{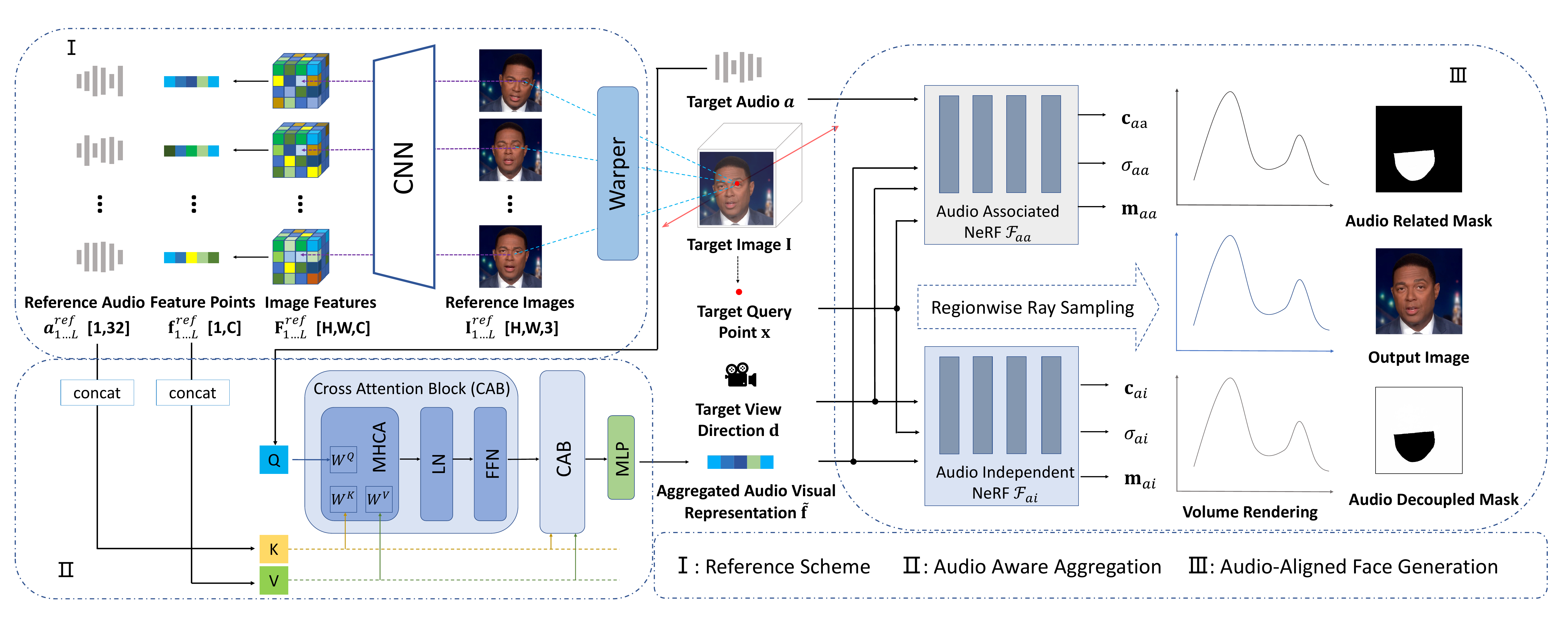} 
\caption{Overview of the proposed \method. The reference scheme gathers the audio visual information from reference images and audio features precisely. \aaa module fuses these features with cross attention and yields a strong representation. \afag strategy models the whole face region with two separated NeRFs, synthesizing the portrait with high fidelity.}
\label{fig_pipeline}
\end{figure*}

On the other hand, we employ a dual-NeRF framework to simultaneously model the \ar and \ad regions of a talking head. 
According to our observations in \cref{fig_mot} (b),
the lower half face can be regarded as the audio related part, whose variations have strong correlations with the audio signal. This part is modeled with an \aac  NeRF that conditions on audio features, while the rest parts are modeled by an \ai NeRF that requires no audio features. 
Thanks to the disentanglement between different face regions and the audio signal, our \textit{\afag} strategy brings better audio-to-lip consistency and finer rendering results.

To summarize, three key contributions are made to improve the practical few shot talking head synthesis.

\begin{itemize}
    \item  We propose an \aaa module based on a reference scheme, which takes full advantage of the audio visual relationships between target and reference images and yields a strong prior. 
    


\item We introduce an \afag strategy to decouple the face modeling into \aac NeRF and \ai NeRF, achieving better audio-lip synchronization and facial details.

  \item Sufficient experiments have proved the superiority of our \method over state-of-the-art on image fidelity, audio-lip synchronization, and generalization ability.  
\end{itemize}

\section{Related Work}
\subsection{Audio-driven Talking Head Generation}
Audio-driven talking head generation aims to animate a speaker according to input audios. 
Image based methods \cite{wav2lip,zakharov2019few} utilize GANs \cite{gan} and Auto-encoders \cite{vae} to generate talking faces with audio signals as conditional inputs. 
Model based methods \cite{atvg,nvp,das2020speech,wang2021one,makeittalk,song2022everybody} leverage structural information such as 2D landmarks or 3DMM parameters for better face modeling. For instance, \cite{atvg} and \cite{nvp} generate faces with predicted facial landmarks or 3DMM expression coefficients. These methods 
can quickly adapt to an unseen identity. However, the prediction error of the representations may lead to inferior image quality, and they usually require hundreds of videos for training. NeRF based methods \cite{adnerf,sspnerf,dfrf} have brought a new trend of talking head synthesis. They perform optimization on the video clip of a single person, and can synthesize pose-controllable faces of any resolution with high fidelity. AD-NeRF \cite{adnerf} use two separated NeRFs to model the head and the torso part respectively. SSP-NeRF \cite{sspnerf} performs rays re-sampling based on the loss magnitude of different semantic regions. 
Despite the above advantages, the generalization ability of NeRF based methods to new identities still needs to be improved, and they suffer from performance drop when the video clip is relatively short.

\subsection{Few Shot Neural Rendering}
Neural Radiance Fields (NeRFs) \cite{mildenhall2020nerf} combines MLPs with differentiable volume rendering and achieves photorealistic view synthesis results. 
Although impressive results are obtained, the original NeRF needs to be retrained for each new scene, which are both time consuming and computational expensive. Moreover, when only sparse views are available, because of the lack of the prior knowledge between scenes \cite{yu2021pixelnerf}, the synthesis results can suffer from a large degradation in quality. 

Few Shot Neural Rendering \cite{grf,yu2021pixelnerf,chen2021mvsnerf,stylenerf,pointnerf} are proposed to alleviate these problems with the assistance of different kinds of priors such as 2D image features \cite{grf,yu2021pixelnerf,ibrnet}, trainable latent codes \cite{codenerf,gafni2021dynamic} and style inputs \cite{stylenerf,eg3d}. 
Among them, the pixel level feature \cite{grf,yu2021pixelnerf,ibrnet} from randomly chose 2D reference images is the most commonly used prior to promote the NeRF's rendering ability when only a few observations are available.  When coming up with a new scene, the NeRF can perform quick generalization based on the reference image features from that scene.
DFRF \cite{dfrf} directly uses the above reference scheme for few shot talking head generation. But it ignores the importance of audio features in talking head rendering. 


\section{Proposed Method}

\subsection{Overview}

The full pipeline of our \method is shown in \cref{fig_pipeline}. Both audio features and visual features are extracted as the references. The proposed \aaa module fuses these features with cross attention and yields a strong prior for fully leveraging the limited data. Then, the audio related and audio decoupled regions are modeled by our \afag strategy. Finally, the portrait of the speaker and semantic masks are synthesized through volume rendering. 

\subsection{NeRF for Audio Driven Talking Head}
The original NeRF encodes a static scene as a continuous volumetric radiance field $\mathcal{F}$, which is modeled by an MLP. It takes a 3D query point $\mathbf{x}$ and its view direction $\mathbf{d}$ as input, and outputs the corresponding density $\sigma$ and color $\mathbf{c}$: $F(\mathbf{x},\mathbf{d}) = (\mathbf{c}, \sigma)$. 
When applying NeRF to talking head, one will take the audio feature as an additional input, and the rendering process can be written as $\mathcal{F}(\mathbf{x},\mathbf{d}, \mathbf{a}) = (\mathbf{c}, \sigma)$. 

\noindent  \textbf{Reference Scheme.}
Despite the superior rendering quality, NeRF-based methods have to optimize each identity individually since no prior knowledge is shared between different identities. 
To improve the generalization ability of NeRFs on few observations, pixel level features from multi-view images (dubbed as reference images) \cite{yu2021pixelnerf,grf} are brought to construct a visual prior. 
DFRF \cite{dfrf} first utilizes this reference scheme in talking head generation, improving rendering ability on few shot datasets to some extent. 
 
Specifically, given $L$ reference images, let $\mathbf{I}_i \in \mathbb{R}^{H_{i} \times W_{i} \times 3}$ and $\mathbf{P}_i \in \mathbb{R}^{3 \times 4}$  denote the $i$-th image and camera projection matrix respectively ($i \in \{0,1,..., L-1\}$). A shallow convolutional network without downsampling is employed to extract dense features $\mathbf{F}_{i} \in \mathbb{R}^{H \times W \times D}$ from each image $\mathbf{I}_i$, where $H$, $W$ and $D$ are the height, width and the feature channel respectively. 
To facilitate the rendering of a 3D point $\mathbf{x}$ on target image,
we first project $\mathbf{x}$ onto $i$-th reference image to obtain image features $\mathbf{f}^{ref}_{i} \in \mathbb{R}^D$. Then all extracted features $\mathbf{f}^{ref}_{1...L}$ are merged as a condition $\tilde{\mathbf{f}}$.
So, an audio-driven NeRF model with reference scheme can be formulated as 
\begin{equation}
\mathcal{F}({\mathbf{x}},\mathbf{d}, \mathbf{a}, \tilde{\mathbf{f}}) = (\mathbf{c}, \sigma).
\end{equation}

In addition, we denote the projection coordinate of the 3D point $\mathbf{x}$ to the i-th 2D reference image as $\mathbf{p}^{ref}_{i} = (\mathbf{u}_{i},\mathbf{v}_{i})$. 
Since the talking head is dynamic, directly performing projection may bring some errors. Thus, an image warper is imposed to calibrate the 2D coordinate by predicting its offset $\Delta \mathbf{p}^{ref}_{i}$ on the feature plane. The calibrated coordinate is denoted by 
\begin{equation}
\mathbf{p}^{{ref}^{\prime}}_{i}=\mathbf{p}^{ref}_{i}+\Delta \mathbf{p}^{ref}_{i}.
\end{equation}

\subsection{\aaa}

In a talking head video, if the speakers in the reference image and the target image have similar speech contents, they tend to have similar mouth shapes, as we mentioned above. Therefore, we introduce \aaa module into the feature fusion process to make the reference image, whose audio is more similar to target, contribute more. Let $\mathbf{a}$, $\mathbf{a}^{ref}_{1...L}$ and $\mathbf{f}^{ref}_{1...L}$ be target audio feature, reference audio feature and reference image feature respectively, then we have


\begin{equation}
\begin{aligned}
 \tilde{\mathbf{f}} = \operatorname{AAA}(\mathbf{a},\mathbf{a}^{ref}_{1...L},\mathbf{f}^{ref}_{1...L}).
\end{aligned}
\end{equation}

The \aaa module utilizes a transformer structure with cross attention blocks to fuse the audio and the visual information. 
To be more concrete, $\mathbf{a}$, $\mathbf{a}^{ref}_{1...L}$, and $\mathbf{f}^{ref}_{1...L}$ are projected and reshaped to get their corresponding tokens $\mathbf{T}_{a}^{tar} \in \mathbb{R}^{N \times 128}$, $\mathbf{T}_{a}^{ref} \in \mathbb{R}^{NL\times 128} $ and $\mathbf{T}_{f}^{ref} \in \mathbb{R}^{NL \times 128}$, $N$ is the number of query point samples in a batch. The audio and image tokens are then modeled by the Multi-Head Cross Attention (MHCA), along with Layer Normalization (LN), Residual Connection (RC) and Feed Forward Network (FFN). The cross attention block, with $\mathbf{T}_{a}^{tar}$ as \emph{query}, $\mathbf{T}_{a}^{ref}$ as \emph{key}, $\mathbf{T}_{f}^{ref}$ as \emph{value}, can be formalized as 
\begin{small}
\begin{equation}
\begin{aligned}
& \operatorname{MHCA}\left(\mathbf{T}_{a}^{tar}, \mathbf{T}_{a}^{ref}, \mathbf{T}_{f}^{ref}\right) = \\ &\operatorname{Softmax}\left[\frac{\mathbf{T}_{a}^{tar} \boldsymbol{W}^{Q}\left(\mathbf{T}_{a}^{ref} \boldsymbol{W}^{K}\right)^{T}}{\sqrt{d}}\right] \mathbf{T}_{f}^{ref} \boldsymbol{W}^{V}, 
\end{aligned}
\end{equation}
\end{small}

\noindent where $\boldsymbol{W}^{Q}\in \mathbf{R}^{128 \times d}$ , $\boldsymbol{W}^{K}\in \mathbf{R}^{128 \times d}$ , $\boldsymbol{W}^{V}\in \mathbf{R}^{128 \times d}$ are the projection matrices with hidden dimension $d$, which is also set to 128. 
It can be seen the closer the \emph{key} (reference audio) and \emph{query} (target audio) are, the greater the weight of \emph{value} (reference feature) in $\tilde{\mathbf{f}}$ will be.
When fitting a new identity, given the reference images and audio features, it can help the NeRF to quickly model the texture and geometry. Two cross attention blocks are involved in the module and their output is passed through two Full Connection layers with a ReLU activation in between, and yields the final aggregated audio visual feature prior $\tilde{\mathbf{f}}$.

We also introduce the audio aware manner into the image warper module. The warper takes the target query point $\mathbf{x}$ and the target audio $\mathbf{a}$ as input, together with the corresponding audio $\mathbf{a}_{ref}$ and ray direction $\mathbf{d}_{ref}$ of the $i$-th reference, and outputs the coordinate offset, achieving more precise feature extraction:
\begin{small}
\begin{equation}
\begin{aligned}
\Delta \mathbf{p}^{ref}_{i}
= (\Delta\mathbf{u}_{i},\Delta\mathbf{v}_{i})
=\operatorname{Warper} (\mathbf{x},\mathbf{a},\mathbf{a}^{ref}_{i},\mathbf{d},\mathbf{d}^{ref}_{i}).
\end{aligned}
\end{equation}
\end{small}

\subsection{\afag }


As stated before, a disentangled modeling of the \ar reigon and the \ad region
is of great significance.
Our \afag strategy uses an \aac NeRF and an \ai NeRF to model these two regions separately,
and only the \aac NeRF conditions on audio feature. 
To merge the rendering results of the two NeRF models, we add 
an additional parsing branch to predict mask $\mathbf{m}_{aa}$ or $\mathbf{m}_{ai}$, where $\mathbf{m}_{aa}$ has 1 in audio related region (i.e., the lower half face ), and 0 in audio independent region, $\mathbf{m}_{ai}$ is otherwise.
Then we have the following two formulations:
\begin{equation}
\begin{aligned}
& \mathcal{F}_{aa}({\mathbf{x}},\mathbf{d}, \mathbf{a}, \tilde{\mathbf{f}}) = (\mathbf{c}_{aa}, \sigma_{aa}, \mathbf{m}_{aa}) \\
& \mathcal{F}_{ai}({\mathbf{x}},\mathbf{d}, \tilde{\mathbf{f}}) = (\mathbf{c}_{ai}, \sigma_{ai}, \mathbf{m}_{ai}).
\end{aligned}
\end{equation}
For a query point $\mathbf{x}$, we can input it into two NeRFs and use the predicted mask to blend the colors or densities of both outputs, like \cite{ma2023semantic}. However, it will double the training time because a batch of rays has to go through the two NeRFs simultaneously.

\noindent \textbf{Regionwise Ray Sampling.} 
Consequently, we elaborate a \rrs mechanism to sample different rays in different sub-regions, to mitigate the computational overhead caused by the dual NeRF. In this mechanism, each NeRF takes as input only the rays from its own corresponding regions and an overlapping region, improving the training speed without damaging the rendering quality.

Concretely, for a set of sampling rays $\Omega$, 
 we use $\epsilon \times \Omega$ to represent a new set that has $\epsilon \times |\Omega|$ rays randomly sampled from $\Omega$ ($|\Omega|$ means the number of rays in $\Omega$). 
 The set of the rays from the \ar region and the \ad region are denoted as $\Omega_{ar}$ and $\Omega_{ad}$. 
An overlap region is defined as  $\Omega_{overlap}=(\epsilon \times \Omega_{ar}) \cup (\epsilon \times \Omega_{ad})$, where rays are fed into two NeRFs simultaneously.
While the remaining parts $\Omega_{aa} = \Omega_{ar} \setminus \Omega_{overlap} $ and $ \Omega_{ai} = \Omega_{ad} \setminus \Omega_{overlap} $ are rendered by their corresponding NeRFs separately. ($\cup$ and $\setminus$ denote sets union and subtraction operations). In practice, $\epsilon$ is set to 0.4 for the best lip generation result. More effects of this \rrs mechanism can be found in the supplementary material. 

\noindent  \textbf{Volume Rendering.} 
 During training, for rays from $\Omega_{aa}$ or $\Omega_{ai}$, the color density and the occupancy are obtained directly from their corresponding NeRFs. For a pixel lying in the overlapping region $\Omega_{overlap}$, its color density and the occupancy become the mixup of the two NeRFs:

\begin{small}
\begin{equation}
\begin{aligned}
& \mathbf{c}  = 
\begin{cases}
\mathbf{m}_{aa} \cdot \mathbf{c}_{aa} + \mathbf{m}_{ai} \cdot \mathbf{c}_{ai}, & \mathbf{r} \in \Omega_{overlap} \\ 
\mathbf{c}_{aa}, & \mathbf{r} \in \Omega_{aa} \\ 
\mathbf{c}_{ai}, & \mathbf{r} \in \Omega_{ai} 
\end{cases} \\
& \sigma  = 
\begin{cases}
\sigma_{aa}+ \sigma_{ai}, & \mathbf{r} \in \Omega_{overlap} \\ 
\sigma_{aa}, &  \mathbf{r} \in \Omega_{aa} \\ 
\sigma_{ai}.  & \mathbf{r} \in \Omega_{ai} 
\end{cases}
\end{aligned}
\end{equation}
\end{small}

During inference, the whole image are regarded as the overlap region, and rendered by two NeRFs at the same time, since there is no ground truth mask available. 
To get the predicted RGB pixel and the mask, we utilize classical volume rendering to accumulate the samples on the ray:
 \begin{small}
 \begin{equation}
 \begin{aligned}
& \hat{\mathbf{C}}(\mathbf{r})=\int_{t_{n}}^{t_{f}} T(t) \sigma(\mathbf{r}(t),\mathbf{a}, \tilde{\mathbf{f}}) \mathbf{c}(\mathbf{r}(t), \mathbf{d}, \mathbf{a}, \tilde{\mathbf{f}}) dt, \\
& \hat{\mathbf{m}}(\mathbf{r}) = \int_{t_{n}}^{t_{f}} T(t) \sigma(\mathbf{r}(t),\mathbf{a}, \tilde{\mathbf{f}}) \mathbf{m}(\mathbf{r}(t), \mathbf{d}, \mathbf{a}, \tilde{\mathbf{f}}) dt, \\
\end{aligned}
\label{eql:C_m}
\end{equation}
\end{small}
 
\noindent where $T(t)=\exp \left(-\int_{t_{n}}^{t} \sigma(s) d s\right)$ denotes for the accumulated transmittance along the ray from $t_{n}$ to $t$, $t_{n}$ and $t_{f}$ are the lower and the upper bound of depth respectively. \cref{eql:C_m} shows the rendering process of the \aac NeRF, where both the audio signal and the aggregated audio visual feature take part in the volume rendering process. For audio independent NeRF, audio signal should be removed.

\subsection{Network Training}
 Following the original NeRF \cite{mildenhall2020nerf}, we use a reconstruction loss term to optimize the coarse and the fine network (we still take \aac NeRF as example if not specified), which can be written as 
\begin{small}
\begin{equation}
\begin{aligned}
& \mathcal{L}_{\mathrm{p}}=\sum_{\mathbf{r}\in \mathcal{R}}\left[\left\|\hat{\mathbf{C}}_{c}\left(\mathbf{r}\right)-\mathbf{C}\left(\mathbf{r}\right)\right\|_{2}^{2}+\left\|\hat{\mathbf{C}}_{f}\left(\mathbf{r}\right)-\mathbf{C}\left(\mathbf{r}\right)\right\|_{2}^{2}\right], \\
& \mathcal{L}_{\mathrm{m}}=\sum_{\mathbf{r}\in \mathcal{R}}\left[\left\|\hat{\mathbf{m}}_{c}\left(\mathbf{r}\right)-\mathbf{m}\left(\mathbf{r}\right)\right\|_{2}^{2}+\left\|\hat{\mathbf{m}}_{f}\left(\mathbf{r}\right)-\mathbf{m}\left(\mathbf{r}\right)\right\|_{2}^{2}\right], \\
\end{aligned}
\end{equation}
\end{small}

\noindent where $\hat{\mathbf{C}}_{c}\left(\mathbf{r}\right)$ and $\hat{\mathbf{C}}_{f}\left(\mathbf{r}\right)$ are the predicted pixels from the coarse and the fine model respectively, $\mathcal{R}$ denotes for a batch of rays, and $\mathbf{C}\left(\mathbf{r}\right)$ is the ground truth pixel color corresponding to each sampled ray. Similarly, $\hat{\mathbf{m}}\left(\mathbf{r}\right)$ and $\mathbf{m}\left(\mathbf{r}\right)$ are the predicted mask and ground truth. For the \aac NeRF, the ground truth mask is defined as 
\begin{small}
\begin{equation}
    \mathbf{m}\left(\mathbf{r}\right) =
    \begin{cases}
      1, & \mathbf{r} \in \Omega_{ar} \\
      0, & \mathbf{r} \in \Omega_{ad}
    \end{cases}
\end{equation}
\end{small}

\noindent while the ground truth mask for the \ai NeRF is the opposite. 
Besides, we use an $l_2$ loss term to regularize the magnitude of the predicted offset of the warper, which can be written as 
\begin{equation}
    \mathcal{L}_{\mathrm{o}}=\frac{1}{L \cdot|\mathcal{P}|} \sum_{i=1}^{L} \sum_{\mathbf{x} \in \mathcal{P}}  \sqrt{\Delta \mathbf{u}_{i}^{2}+\Delta \mathbf{v}_{i}^{2}},
\end{equation}
where $\mathcal{P}$ is the collection of all the 3D query points. 

Our final loss term can be given as
\begin{equation}
    \mathcal{L}=\mathcal{L}_{\mathrm{p}} + \lambda_{m}\mathcal{L}_{\mathrm{m}} + \lambda_{o}\mathcal{L}_{\mathrm{o}},
\end{equation}
where $\lambda_{m}$ and $\lambda_{o}$ are weight parameters. 

\begin{figure*}[t]
\begin{center}
\includegraphics[width=0.95\linewidth]{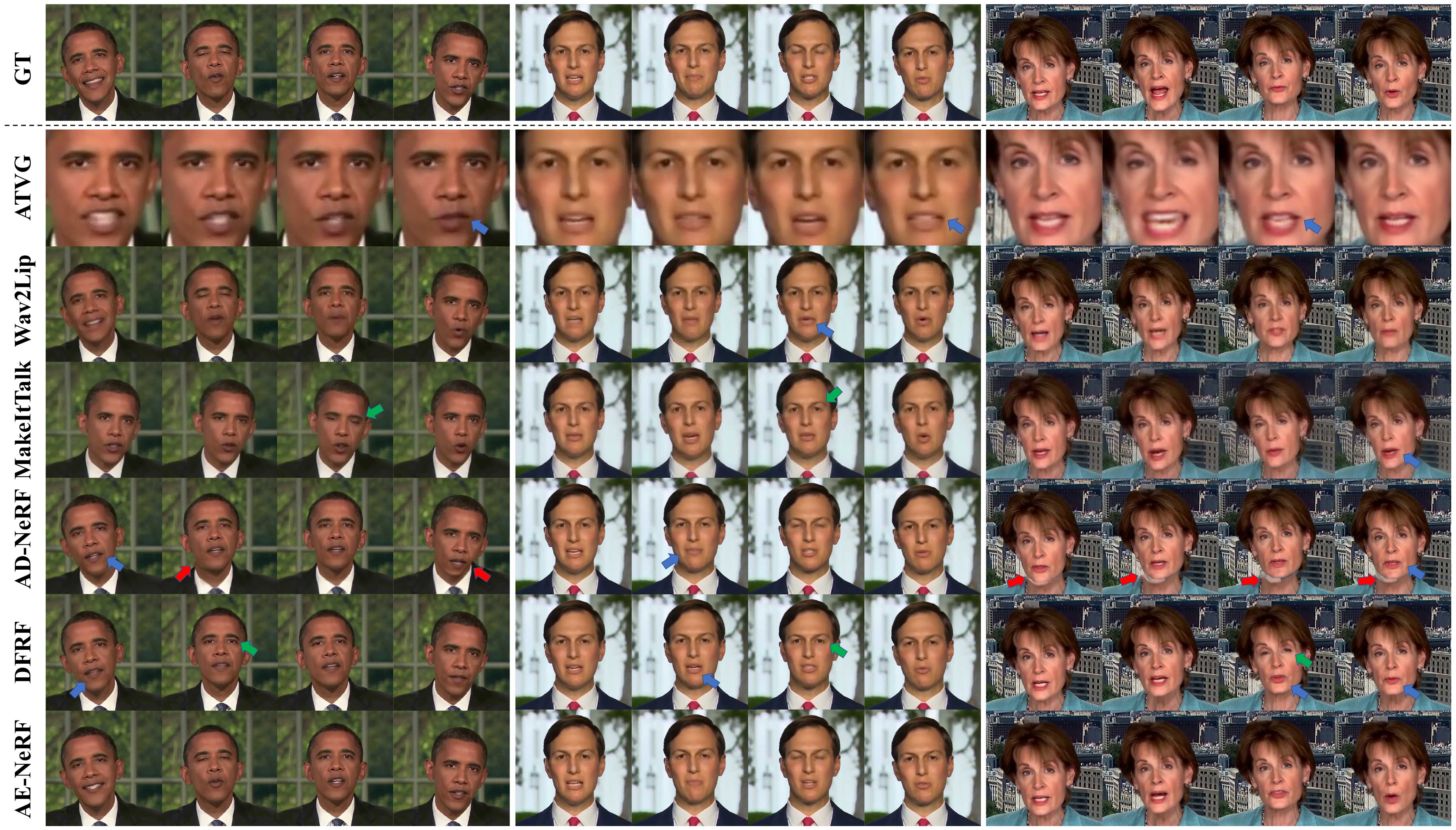}
\end{center}
   \caption{Qualitative comparison with other baseline methods for generated key frame results. We use the blue arrow to denote the inaccurate lip synthesis results like incorrect shape or blurred lips, the red arrow to denote the head-torso inconsistency, and the green arrow to represent inaccurate expression synthesis results. }
\label{fig_quality}
\end{figure*}

\section{Experiments}

\subsection{Experimental Setup}

\textbf{Dataset Preparation.} 
We use the videos provided by AD-NeRF \cite{adnerf}, DFRF \cite{dfrf}, and HDTF dataset \cite{zhang2021flow} to conduct our experiments. Videos are all resampled to 25 fps and resized to a resolution of $512\times512$. For each video, the first half of it is used for training and the second half is used for inference. 


\noindent
\textbf{Baseline Methods.}
We compare our method with one image based method Wav2Lip \cite{wav2lip}, two model based methods ATVG \cite{atvg} and MakeitTalk \cite{makeittalk}, and two NeRF based methods AD-NeRF \cite{adnerf} and DFRF \cite{dfrf}. For the first three methods, we use their official code and provided pretrained models. For AD-NeRF and DFRF, we retrain them on each video on the same number of iterations as our method for fair comparison. DFRF also has a base model like our method, and we have tried to pre-train a base model for AD-NeRF like DFRF and our method, but it fails to generate plausible results for the lack of generalization ability. Comparison with another SOTA NeRF-based method SSP-NeRF \cite{sspnerf} whose training code is not provided, more implementation details, the limitation of our method and the ethical consideration can be found in the supplementary material. 

\noindent 
\textbf{Evaluation Metrics.}
We employ evaluation metrics that have been previously used in talking head generation. We use PSNR and SSIM  to evaluate the image level quality of generated results, LPIPS \cite{lpips} to evaluate the feature level quality. We also use Landmark Distance (LMD) \cite{lmd} and SyncNet Confidence \cite{syncnet} to further measure the mouth shapes and the audio visual synchronization.

\begin{table}[t]\large
\setlength\tabcolsep{3pt}
\begin{center}
\scalebox{0.7}{
\begin{tabular}{@{}c|ccccc|c|c@{}}
\toprule
           & \multicolumn{5}{c|}{Testset A}                        & Testset B & Testset C \\ \midrule
Methods    & PSNR  $\uparrow$                & SSIM $\uparrow$ & LPIPS $\downarrow$ & LMD  $\downarrow$ & Sync $\uparrow$ & Sync $\uparrow$     & Sync $\uparrow$     \\ \midrule
gt         & $\infty$ & 1     & 0     & 0     & 8.545 & 8.406     & 8.873     \\ \midrule
ATVG       & 19.12                 & 0.646 & 0.523 & 2.591  & 5.657 & 4.726     & 6.315     \\
Wav2Lip    & 29.64                 & 0.843 & 0.423 & 2.612 & \textbf{9.750}  & \textbf{7.824}     & \textbf{10.715}    \\
Makeittalk & 22.28                 & 0.655 & 0.480  & 10.720 & 5.945 & 4.378     & 5.556     \\
AD-NeRF    & 27.73                & \underline{0.881} & 0.202 & \underline{2.603} & 4.274 & 4.230      & 4.656     \\
DFRF       & \underline{32.30}                  & \textbf{0.949} & \underline{0.080}  & 3.023 & 5.219 & 4.859     & 5.321     \\
\method       & \textbf{32.63}                 & \textbf{0.949} & \textbf{0.078} & \textbf{2.425}  & \underline{6.904} & \underline{6.217}     & \underline{6.690}      \\ \bottomrule
\end{tabular}
}
\end{center}
\caption{Method comparison under self-driven (Testset A) and cross-driven (Testset B and C) setting. The best and the second results are in bold and underlined respectively.}
\label{tab_cmp}
\end{table}

\subsection{Face Quality Comparison}
To compare the quality of the generated talking head thoroughly, two different settings are taken into account: Self driven setting, where the video and the audio are from the same person. Cross driven setting, where the audio from one person is used to drive another identity. Each video is about two minutes in length.

\begin{figure}[t]
\begin{center}
\includegraphics[width=0.99\linewidth]{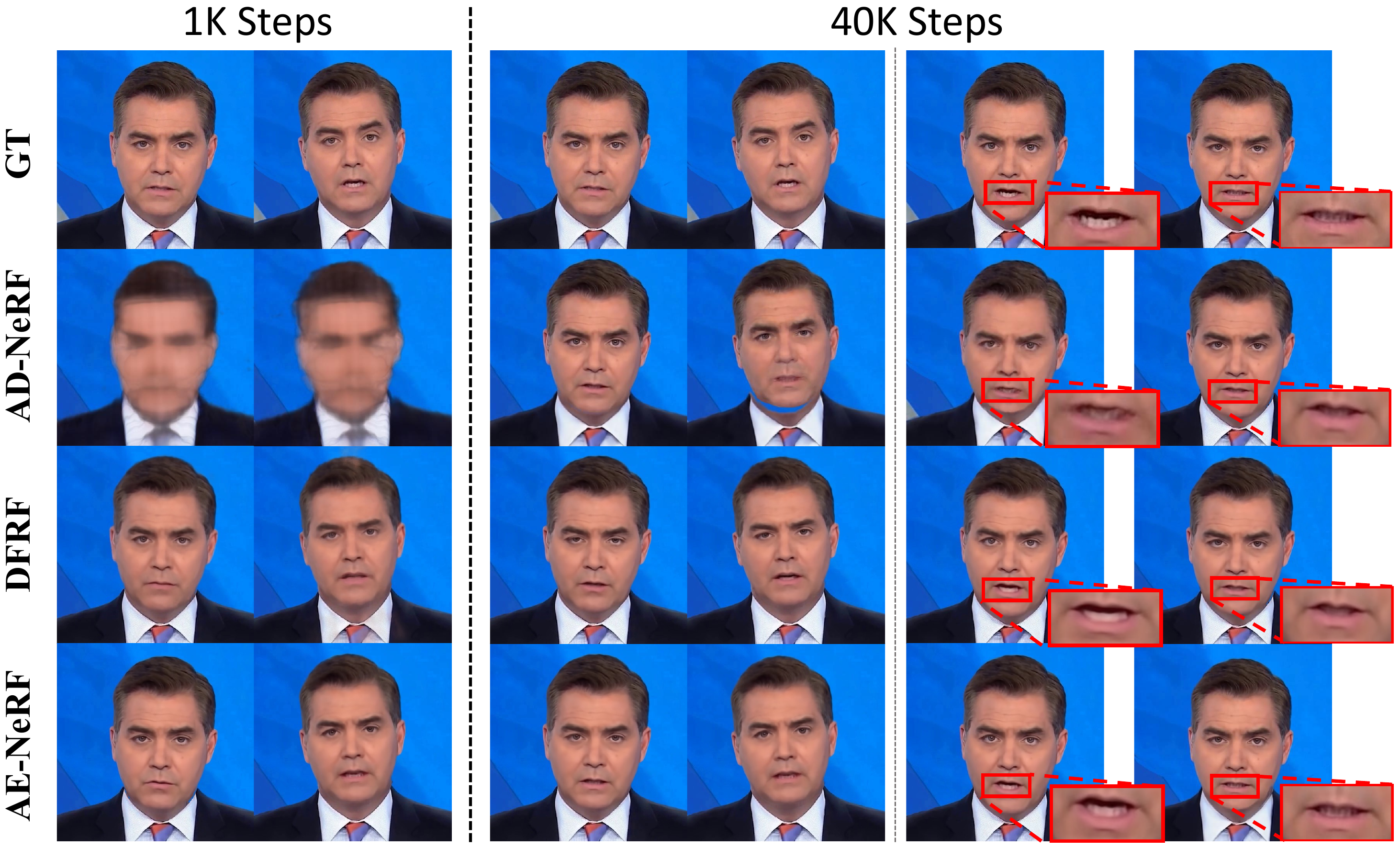}
\end{center}
   \caption{Qualitative comparison against other methods with 1k and 40k finetuning steps. 15s video clip is available for training.}
\label{fig_ft_steps}
\end{figure}

\noindent \textbf{Results under Self-driven Setting.} Key frames generated according to the Obama video in \cite{adnerf} and the broadcaster videos in \cite{dfrf} are shown in Fig.\ref{fig_quality}. 
NeRF based methods have shown superiority image quality against image based methods and model based methods, and have managed to generate high fidelity synthesis results. But AD-NeRF suffers from head-torso misalignment as the red arrows pointed out, while DFRF fails to generate some face details correctly. Besides, both AD-NeRF and DFRF tend to generate lips misaligned from the ground truth. Our \method has shown the best lip-alignment with the ground truth frames, as well as the facial details.

Quantitative comparison results on AD-NeRF and DFRF videos are shown in Testset A part of Tab.\ref{tab_cmp}. Wav2Lip \cite{wav2lip} uses a pretrained SyncNet \cite{syncnet} as the optimization objective and it achieves a SyncNet score even better than the ground truth. However, the quality of its generated images is relatively low.  NeRF-based methods have shown their superiority not only at the pixel level (PSNR and SSIM) but also at the feature level (LPIPS). AD-NeRF performs slightly worse than DFRF and our method in image quality metrics due to the head-torso misalignment. 
It can be seen that our \method surpasses baseline methods by a large margin in SyncNet confidence and LMD, which indicates the effect of learning aggregated audio visual features for lip synthesis.

\begin{table}[t]\Large
\setlength\tabcolsep{2pt}
\begin{center}
\scalebox{0.7}{
\begin{tabular}{@{}c|cccccc@{}}
\toprule
Method                   & Length & PSNR  $\uparrow$  & SSIM $\uparrow$ & LPIPS $\downarrow$ & LMD  $\downarrow$ & Sync $\uparrow$ \\ \midrule
\multirow{3}{*}{AD-NeRF} & 10s    & 22.83 & 0.846 & 0.156 & 2.427 & 1.587   \\
                         & 15s    & 23.37 & 0.867 & 0.138 & 2.116 & 3.528   \\
                         & 20s    & 23.01 & 0.856 & 0.142 & 1.684 & 4.102   \\ \midrule
\multirow{3}{*}{DFRF}    & 10s    & 28.87 & 0.926 & 0.076 & 1.971 & 3.512   \\
                         & 15s    & 29.60  & 0.938 & 0.066 & 1.804 & 3.688   \\
                         & 20s    & 31.32 & 0.942 & 0.069 & 1.84  & 4.459   \\ \midrule
\multirow{3}{*}{\method}    & 10s    & 28.93 & 0.930  & 0.072 & 1.944 & 6.102   \\
                         & 15s    & 29.52 & 0.938 & 0.067 & 1.766 & 6.217   \\
                         & 20s    & \textbf{31.49} & \textbf{0.946} & \textbf{0.064} & \textbf{1.528} & \textbf{6.743}   \\ \bottomrule
\end{tabular}}
\end{center}
\caption{Method comparison with different training data length under 40k iterations.}
\label{tab_ft_len}
\end{table}

\noindent \textbf{Results under Cross-driven Setting.} 
Cross driven results are shown in Testset B and Testset C of Tab.\ref{tab_cmp}. Audios from HDTF dataset are used to drive other identities. We only calculate the SyncNet score since there is no ground truth for other metrics.  AD-NeRF and DFRF fail to synthesize accurate lip shapes according to audios from different speakers. We attribute this to the lack of aggregated audio visual information. Our \method have shown competitive performance in audio-lip consistency.

\subsection{Few Shot Talking Head Synthesis}
We compare our \method with other NeRF based talking head methods under a more challenging setting, few shot talking head synthesis, which further validate the generation ability of our method.

\noindent \textbf{Synthesizing Talking Head with Short Videos.}
Firstly, we show the performance of different NeRF based methods on very short videos, with each model from a different method being fine-tuned by 40k steps. Metrics calculated with training data lengths of 10s, 15s, and 20s are shown in Tab. \ref{tab_ft_len}. It can be seen that when training under a few shot setting, our method still maintains high image generation quality and audio visual alignment. 
Comparison of different faical details are shown in Fig. \ref{fig_ft_steps}. Our method keeps the best of the original facial details, especially at the mouth region, while AD-NeRF and DFRF sometimes generate incorrect lip shape. We attribute this phenomenon to the lack of feature prior and the naive visual feature fusion process without considering the audio information. This further validates the face modeling ability of our \method.

\begin{figure}[t]
\begin{center}
\includegraphics[width=1.0\linewidth]{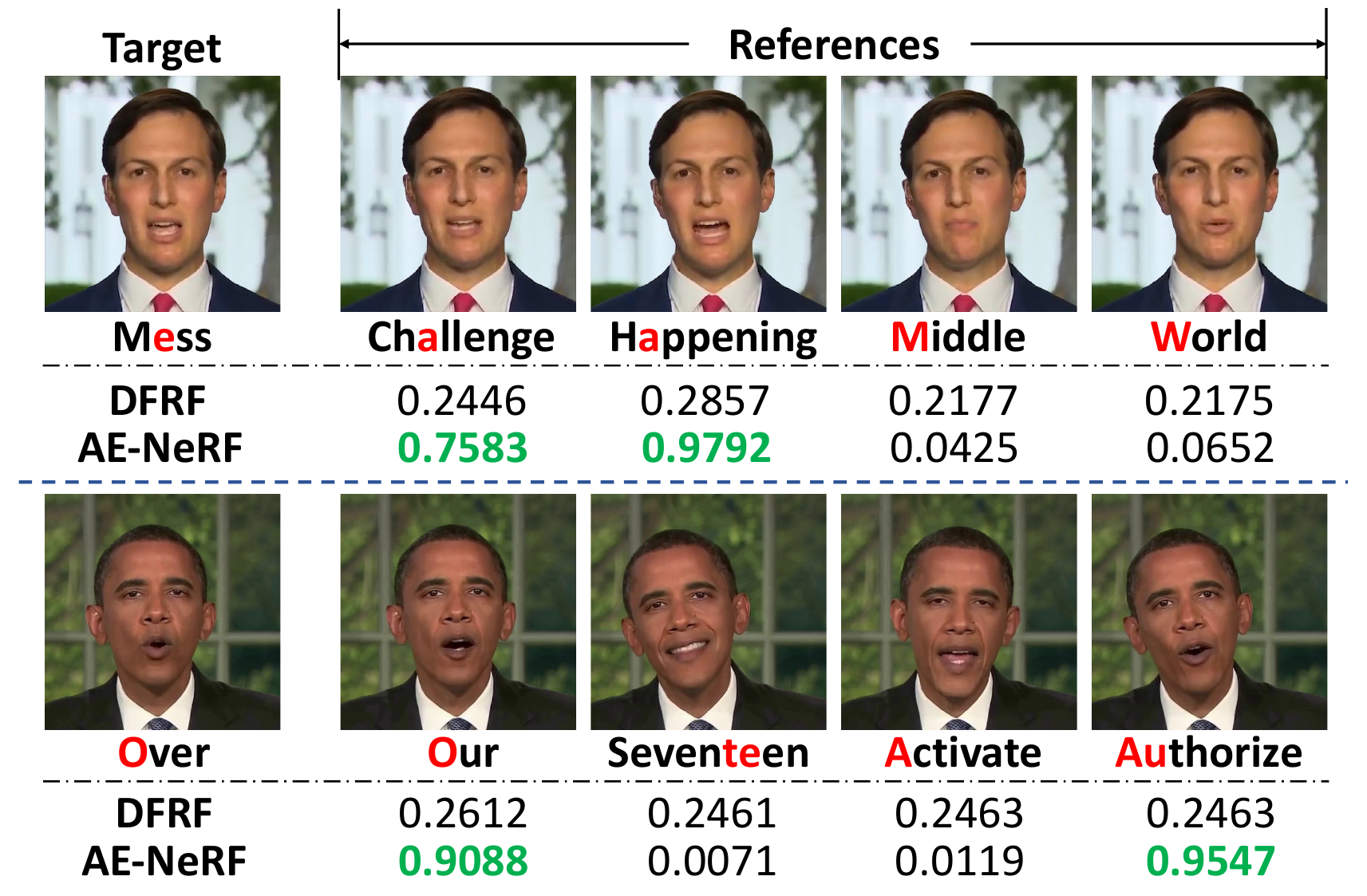}
\end{center}
   \caption{Attention score between the target feature and different reference features in the cross attention block. Similar audio visual contents bring higher attention scores.}
\label{fig_attn}
\end{figure}

\begin{table}[t]\large
\setlength\tabcolsep{2pt}
\begin{center}
\scalebox{0.85}{
\begin{tabular}{@{}cc|ccccc@{}}
\toprule
\multicolumn{2}{c|}{Methods}                        & PSNR $\uparrow$ & SSIM $\uparrow$ & LPIPS $\downarrow$ & LMD $\downarrow$ & Sync $\uparrow$ \\ \midrule
\multicolumn{2}{c|}{ground truth}                   & $\infty$        & 1               & 0                  & 0                & 8.065           \\ \midrule
\multicolumn{1}{c|}{\multirow{3}{*}{AD-NeRF}} & 1k  & 22.13           & 0.777           & 0.305              & 9.158            & 0.291           \\
\multicolumn{1}{c|}{}                         & 10k & 23.72           & 0.868           & 0.143              & 2.148            & 3.697           \\
\multicolumn{1}{c|}{}                         & 40k & 23.37           & 0.867           & 0.138              & 2.116            & 3.528           \\ \midrule
\multicolumn{1}{c|}{\multirow{3}{*}{DFRF}}    & 1k  & 29.39           & 0.931           & 0.086              & 1.918            & 2.995           \\
\multicolumn{1}{c|}{}                         & 10k & 29.47           & 0.936           & 0.073              & 1.905            & 2.985           \\
\multicolumn{1}{c|}{}                         & 40k & \textbf{29.60}            & 0.936           & 0.070              & 1.804            & 3.688           \\ \midrule
\multicolumn{1}{c|}{\multirow{3}{*}{\method}}    & 1k  & 29.18           & 0.930            & 0.087              & 1.893            & 5.552            \\
\multicolumn{1}{c|}{}                         & 10k & 29.32           & 0.937           & 0.072              & \textbf{1.765}            & 6.044            \\
\multicolumn{1}{c|}{}                         & 40k & 29.52           & \textbf{0.938}           & \textbf{0.067}              & 1.766            & \textbf{6.217}           \\ \bottomrule
\end{tabular}}
\end{center}
\caption{Method comparison with 15s training clip under different training iterations.}
\label{tab_fttime}
\end{table}

\noindent \textbf{Synthesizing Talking Head with Few Iterations. }
We further explore the generation ability of different methods with few training iterations. 
We utilize video clips released from DFRF \cite{dfrf} to carry out our experiment. Each video is 30s in length. 
We compare the portraits generated after 1k and 40k training steps to further show the synthesis results of different methods, which is also shown in Fig. \ref{fig_ft_steps}. Within 1k steps, DFRF and our method can fit the new identity, which AD-NeRF fails to generalize on. In 40k steps, our method can generate portraits with higher fidelity than DFRF. 
Quantitatively, results with different training steps are shown in Tab. \ref{tab_fttime}, where our \method achieves similar image quality metrics with DFRF under the same training step, but it has shown superiority in SyncNet confidence and LMD, indicating better lip synthesis results.

\begin{figure}[t]
\begin{center}
\includegraphics[width=0.95\linewidth]{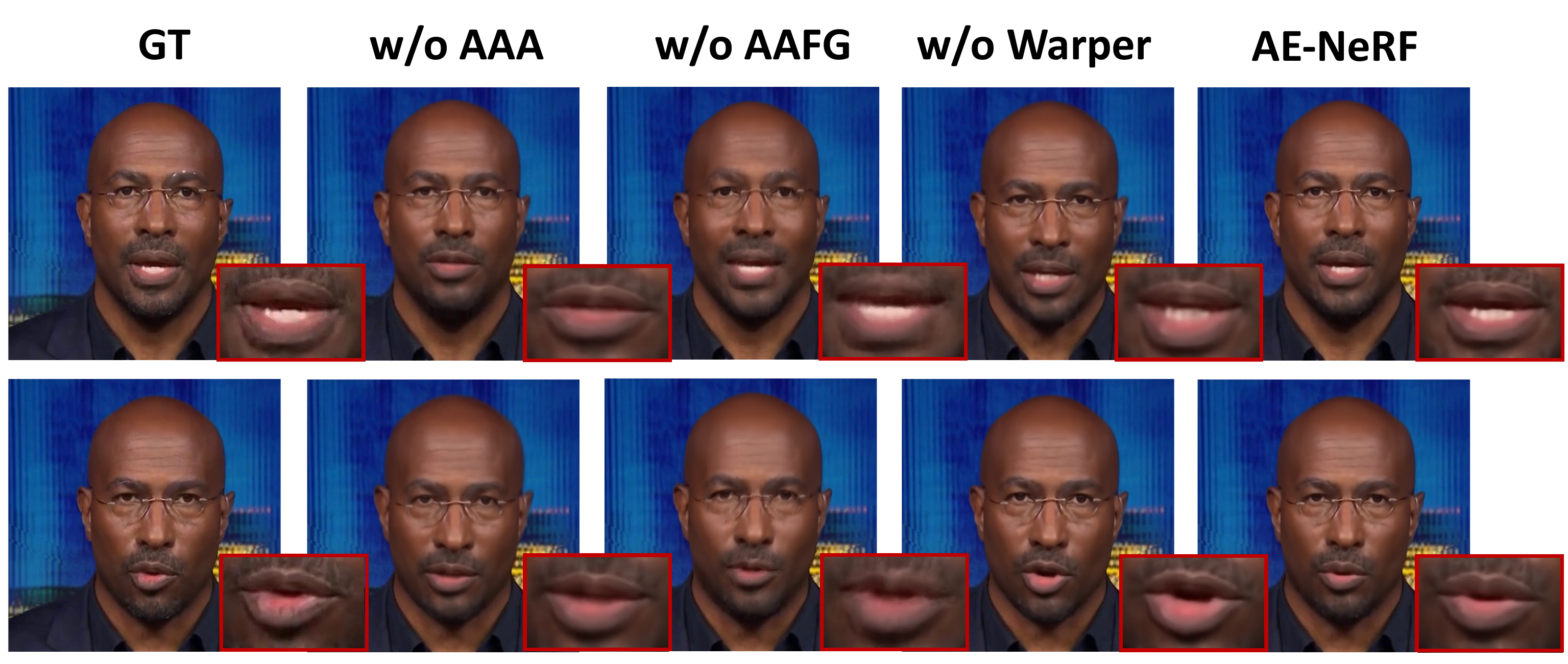}
\end{center}
   \caption{Qualitative ablation study}
\label{fig_ablation}
\end{figure}

\begin{table}[t]\Large
\begin{center}
\setlength\tabcolsep{2pt}
\scalebox{0.8}{
\begin{tabular}{@{}cccccc@{}}
\toprule
Methods          & PSNR $\uparrow$     & SSIM $\uparrow$  & LPIPS $\downarrow$ & LMD $\downarrow$  & Sync $\uparrow$  \\ \midrule
GT               & $\infty$ & 1     & 0     & 0     & 8.545 \\ \midrule
w/o AAA         & 31.25    & 0.929 & 0.092 & 3.646 & 4.570 \\
w/o AAFG         & 32.30    & 0.940 & 0.080 & 3.938 & 6.282 \\
w/o Warper & 32.49    & 0.949 & 0.076 & 2.779 & 6.952 \\
\method         & \textbf{32.64}    & \textbf{0.950} & \textbf{0.076} & \textbf{2.613} & \textbf{7.813} \\ \bottomrule
\end{tabular}
}
\end{center}
\caption{Quantitative ablation study on different modules.}
\label{table_ablation}
\end{table}

\subsection{Ablation Study}

An ablation study is conducted to show the function of each proposed module. 
We replace our \aaa with a slot attention module \cite{slotattention} which simply fuses the visual feature without considering audio signals, denoted as w/o AAA. We also test the performance of the model without \afag, where an \aac NeRF is used to model the whole face area, denoted as w/o AAFG. 
The model trained without the warper is denoted as w/o Warper.  
The qualitative and quantitative results are shown in Fig. \ref{fig_ablation} and Tab. \ref{table_ablation}.
The \aaa module brings better quality in both pixel and feature level. Model obtained without this module has less awareness of the audio signal, resulting in inferior image quality. 
Our \afag strategy brings more correct lip shapes and more natural expressions. The warper can bring more accurate visual feature points, which can also affect the audio lip synchronization.


To further study the effect of the \aaa module, we calculate the average attention scores between different references and the target feature in the first attention block of our \aaa module, shown in Fig. \ref{fig_attn}. As a comparison, scores in the feature aggregation module used in DFRF are also taken into account. The inner product between target-reference pairs with similar audio visual contents is significantly higher than those with distinct contents in our cross attention block. 
This proves that the audio visual interaction process assists the rendering of the target pixel, resulting in better generation results. 


\section{Conclusion}
This work presents Audio Enhanced Neural Radiance Field (\method) for few shot talking head synthesis. Our method consists of an \aaa module which learns a strong prior for improving the generalization ability and an \afag strategy to better model the \ar and the \ad face regions. Comparisons between SOTA methods confirm that our \method  achieves better image quality and fidelity under the custom scenario and a few shot setting.

\section{Ethical Statement}
Our \method is capable of generating vivid speech portraits with high fidelity, 
and can be applied to various situations such as virtual human, 
digital games and film making. 
On the other hand, the misuse of the talking head synthesis technique can lead to moral and legal issues, 
such as crafting malicious DeepFake videos. 
We are committed to fighting against this kind of egregious behavior and use our code and models in the development of the DeepFake detection models. 

\section{Acknowledgments}
This work is supported by the National Key Research and Development Program of China under Grant No. 2021YFC3320103, 
the National Natural Science Foundation of China (NSFC) under Grants 62372452, U19B2038,
and by Alibaba Group through Alibaba Innovative Research Program.
\bibliography{aaai24.bib}   
\clearpage
\section{Implementation Details}
The number of the reference images $L$ is set to 4, and loss weights $\lambda_{m}$ and $\lambda_{o}$ to 1e-3 and 5e-8 respectively. The proportion of the overlap region $\epsilon$ is set to 0.4. 
We use PyTorch to implement the whole framework. All the experiments are conducted on one NVIDIA A100 GPU. Positional encoding is applied to all the 3D query points $\mathbf{x}$ and the view directions $\mathbf{d}$. 
Wav2Vec \cite{baevski2020wav2vec} is utilized to extract the audio feature $\mathbf{a}$ with temporal smoothing.
We use face2face \cite{face2face} to estimate pose matrix $\mathbf{P}$ like previous works \cite{adnerf,dfrf}. 
Adam solver with a learning rate of 5e-4 with exponential learning rate decay is used to train the whole model end to end.  It takes about 36 hours to train a base model and 3 hours for finetuning on a new identity.

\section{Implementation of the Warper} 
\begin{figure}[t]
\begin{center}
\includegraphics[width=1\linewidth]{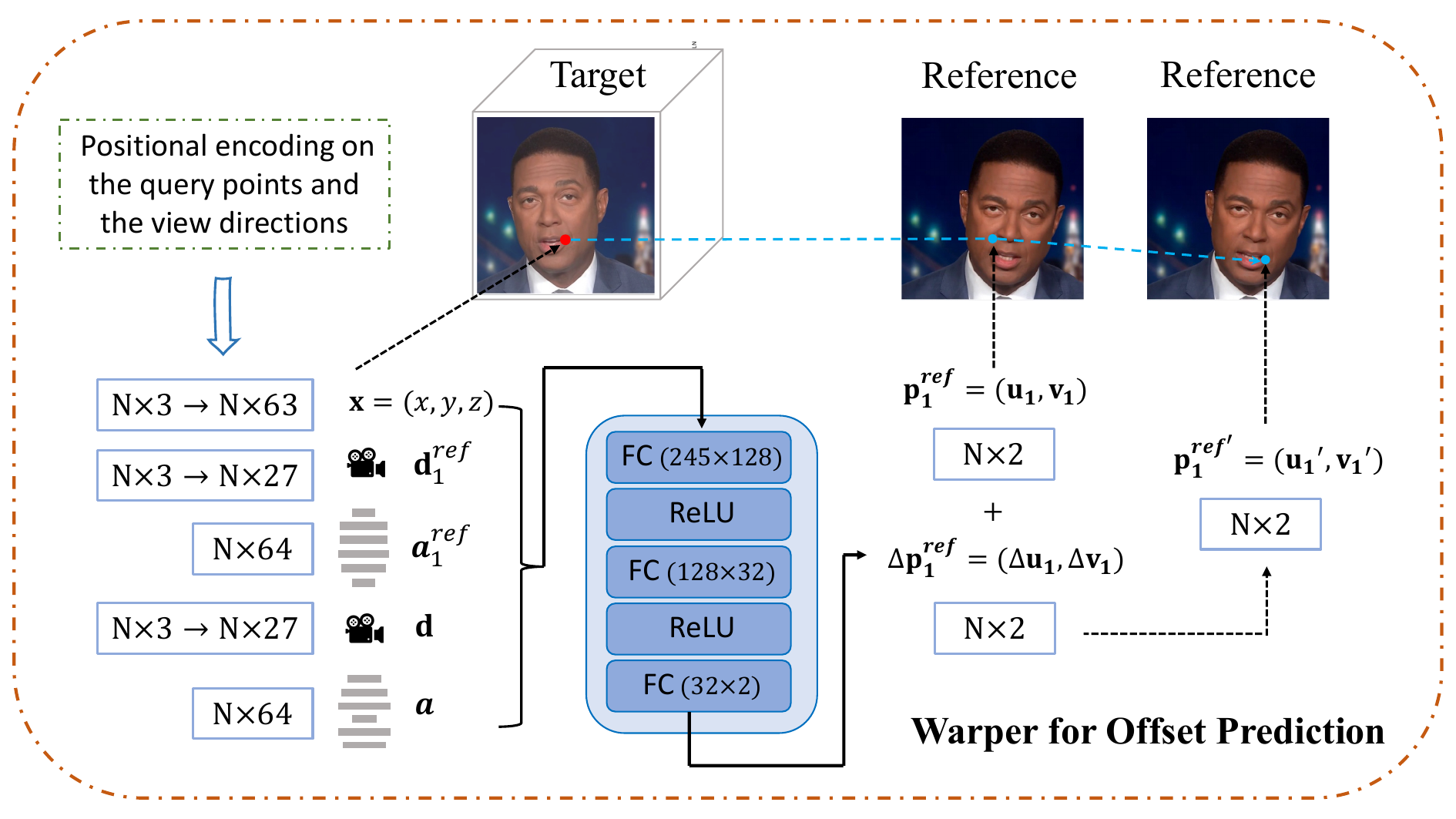}
\end{center}
   \caption{Implementation details of the warper.}
\label{fig_warper}
\end{figure}

The warper is implemented with three Full Connection layers (FC) with ReLU activation in between. It takes the query 3D point $\mathbf{x}$, the target audio $\mathbf{a}$ and view direction $\mathbf{d}$, the $i-$th reference audio $\mathbf{a}^{ref}_{i}$,  as well as $i-$th reference view direction $\mathbf{d}^{ref}_{i}$ as inputs and predicts the offset $\Delta \mathbf{p}^{ref}_{i}$ for more accurate feature extraction. This is shown in \cref{fig_warper}.

\section{Details of \afag}
\begin{figure*}[ht]
\begin{center}
\includegraphics[width=0.95\linewidth]{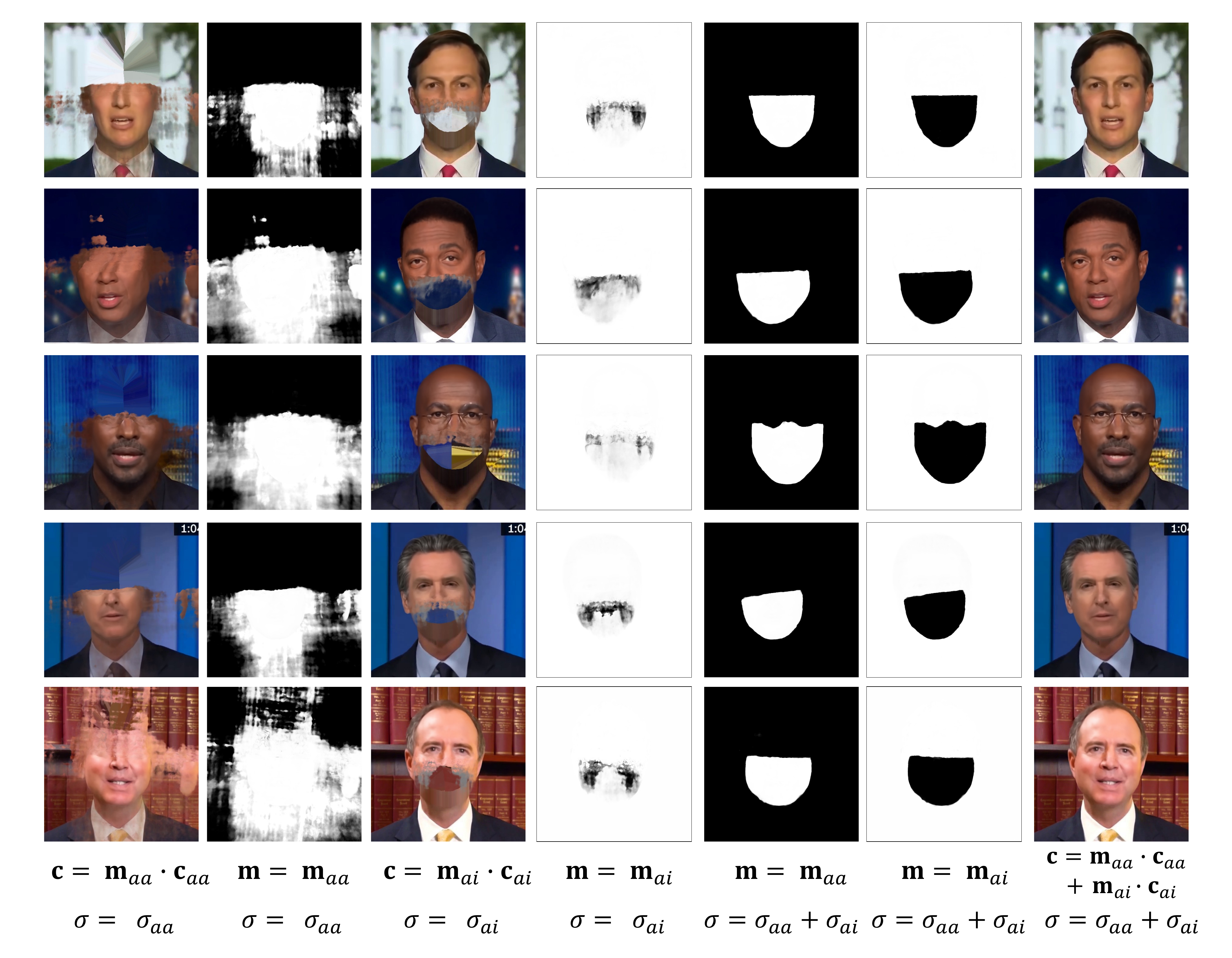}
\end{center}
   \caption{Generated faces of each individual NeRF and the whole dual NeRF framework.}
\label{fig_dual}
\end{figure*}

\textbf{Visualization of the Dual NeRFs.}
The synthesized face, together with the mask generated by the \aac NeRF and the \ai NeRF, and the whole dual NeRF framework are shown in \cref{fig_dual}. 

\noindent \textbf{Effect of \rrs.} 
The effect of \rrs mechanism are shown in \cref{tab_rrs}. The model trained without \rrs is denoted as w/o RRS, where all the rays have to forward though the \aac NeRF and the \ai NeRF. A model trained without \afag is also taken into consideration, where the whole face area is modeled by an \aac NeRF. Our \rrs results in a significant speed improvement, reducing the training time by nearly 10 hours. 
We set $\epsilon$ to 0.4, since it brings the best lip synthesis quality, meanwhile keeps an acceptable training speed. Lastly, although DFRF and w/o AAFG model are faster, they both achieve inferior face quality than the full \method model.

\section{Comparison with SSP-NeRF}
\begin{figure}[t]
\begin{center}
\includegraphics[width=0.95\linewidth]{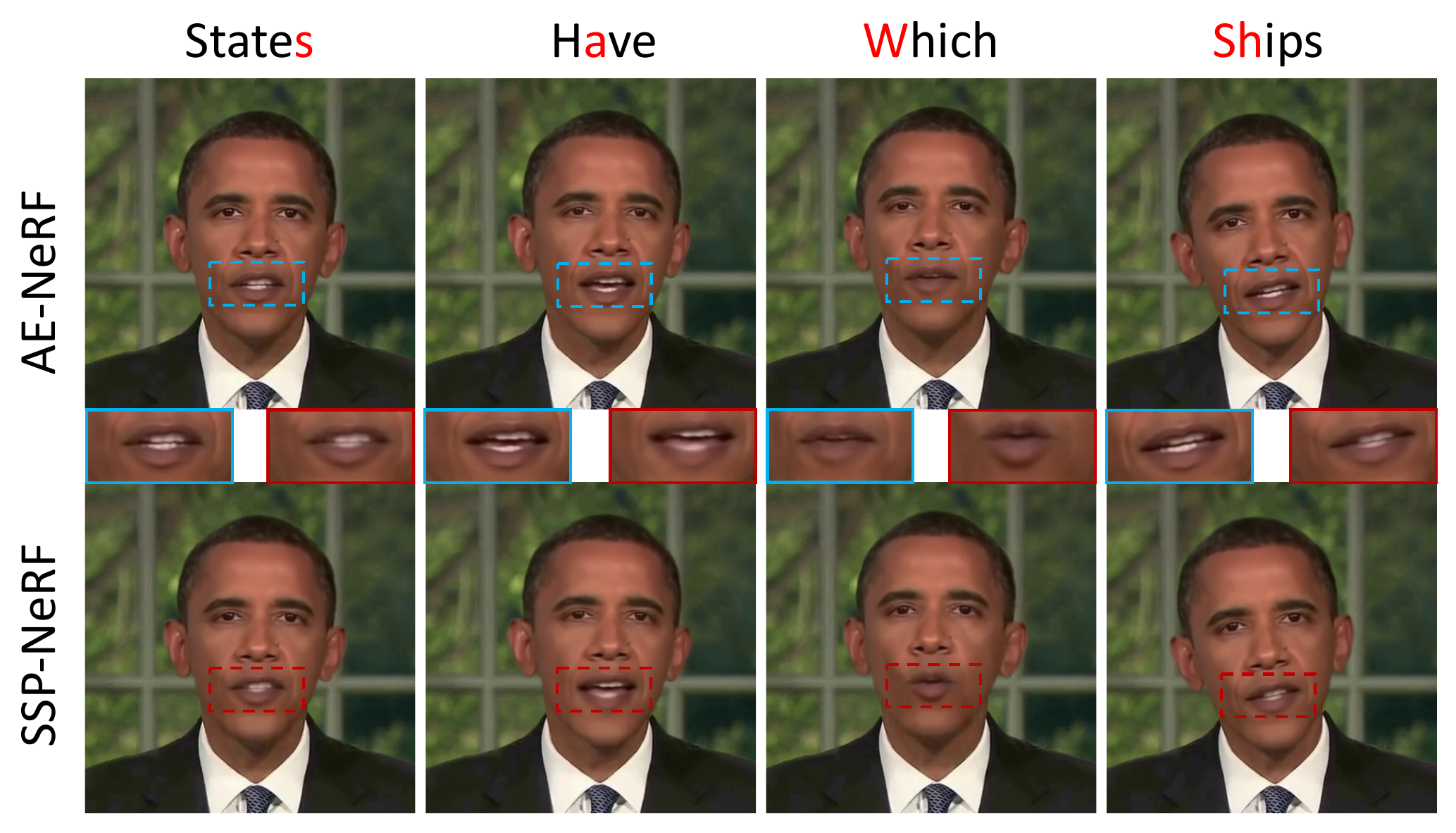}
\end{center}
   \caption{Qualitative comparison with SSP-NeRF. Our AE-NeRF achieves better lip synthesis results.}
\label{fig_ssp}
\end{figure}

Since the code and the model of SSP-NeRF \cite{sspnerf} is not provided. We perform a qualitative comparison using the demo video it provided. See in \cref{fig_ssp}. Our \method synthesizes better lip shape and clearer teeth.

\begin{table}[t]
\setlength\tabcolsep{1.5pt}
\begin{center}
\scalebox{0.85}{
\begin{tabular}{@{}ccccccc@{}}
\toprule
Method           & SSIM $\uparrow$ & PSNR $\uparrow$ & LPIPS $\downarrow$ & LMD $\downarrow$ & Sync $\uparrow$ & \thead{Training \\Time / h} $\downarrow$ \\ \midrule
Ground Truth     & 1               & $\infty$        & 0                  & 0                & 8.968           &   -                \\ \midrule
DFRF             & 0.961           & 29.15           & 0.060               & 2.516            & 3.598           & 36.8              \\
$\epsilon$ = 0.2 & \textbf{0.962}           & 29.48           & 0.060               & 2.001            & 7.268           & \textbf{\color{blue}{40.5}}             \\
$\epsilon$ = 0.4  & 0.961           & 29.31           & 0.061              & 2.029            & \textbf{7.775}           & \textbf{\color{blue}{43.1}}                \\
$\epsilon$ = 0.6  & 0.961           & 29.22           & 0.061              & 1.997            & 6.991           & \textbf{\color{blue}{44.9}}             \\
w/o RRS          & \textbf{0.962}           & \textbf{29.54}           & \textbf{0.058}              & \textbf{1.948}            & 7.611           & \textbf{\color{red}{50.6}}             \\
w/o AAFG         & 0.961          & 29.36           & 0.059              & 2.233            & 6.182           & 31.8            \\ \bottomrule
\end{tabular}
}
\end{center}
\caption{Effect of \rrs.}
\label{tab_rrs}
\end{table}

\section{Video Generation Results}
Please refer to our supplementary video for a direct comparison of different methods.

\section{Limitations and Future Work}
\method uses two NeRFs to model different areas of the face separately, dramatically improving the quality of the rendered talking head with limited data at the expense of some speed. Further, we will introduce existing NeRF acceleration methods such as TensorRF \cite{tensorf} and Instant-NGP \cite{instantngp} to improve the speed of the entire model. We will keep on improving our \rrs strategy for faster inference, e.g. designing a learnable module for more efficient ray allocation. In our future work, we will devote to improving the generalization of NeRF-based talking heads on datasets with extreme poses and expressions.

\section{Ethical Consideration}
Our \method is capable of generating vivid speech portraits with high fidelity, and can be applied to various situations such as virtual human, digital games and film making. On the other hand, the misuse of the talking head synthesis technique can lead to moral and legal issues, such as crafting malicious DeepFake videos. We are committed to fighting against this kind of egregious behavior and use our code and models in the development of the DeepFake detection models. 
\end{document}